\documentclass[letterpaper, 10 pt, conference]{ieeeconf}
\IEEEoverridecommandlockouts
\makeatletter
\let\NAT@parse\undefined
\makeatother
\usepackage{hyperref}
\usepackage[utf8]{inputenc}
\usepackage{authblk}
\usepackage[ruled, linesnumbered]{algorithm2e}
\usepackage{algorithmic}
\usepackage{marginnote}
\usepackage{amsfonts}
\usepackage{amsmath}
\usepackage{graphicx}
\usepackage{subcaption}
\usepackage[table,xcdraw,dvipsnames]{xcolor}

\newtheorem{definition}{Definition}

\usepackage{cite}
\usepackage[nameinlink]{cleveref}
\title{Evaluating Guiding Spaces for Motion Planning}

\author{Amnon Attali$^{1}$, Stav Ashur$^{1}$, Isaac Burton Love$^{1}$, Courtney McBeth$^{1}$, James Motes$^{1}$, Diane Uwacu$^{2}$,\\ Marco Morales$^{1}$, Nancy M. Amato$^{1}$%
\thanks{$^{1}$Department of Computer Science, University of Illinois, 201 N. Goodwin Avenue, Urbana, IL 61801, USA}%
\thanks{$^{2}$Department of Computer Science and Engineering, Texas A\&M University, College Station, Texas, 77843-3112, USA}%
}

\date{September 2022}

\newcommand{\C}{\mathcal{C}}
\newcommand{\GS}{\mathcal{S}}
\begin{document}

\maketitle

\begin{abstract}
    Randomized sampling based algorithms are widely used in robot motion planning due to the problem's intractability, and are experimentally effective on a wide range of problem instances. Most variants do not sample uniformly at random, and instead bias their sampling using various heuristics for determining which samples will provide more information, or are more likely to participate in the final solution. In this work, we define the \emph{motion planning guiding space}, which encapsulates many seemingly distinct prior works under the same framework. In addition, we suggest an information theoretic method to evaluate guided planning which places the focus on the quality of the resulting biased sampling. Finally, we analyze several motion planning algorithms in order to demonstrate the applicability of our definition and its evaluation.
\end{abstract}

\section{Introduction}

An instance of a motion planning problem involves computing a collision-free path between configurations of a robot in a continuous space. The family of sampling based motion planning (SBMP) algorithms uses random sampling of configuration space, and local attempts to connect valid configurations, in order to find such a path \cite{lavalle1998rapidly, KavrakiSLO96}. SBMP is a powerfully general approach, which assumes very little prior knowledge of the underlying space. On the other hand, real world robotics problems contain much underlying structure. In particular, realistic configuration spaces are not arbitrary topological spaces, but rather arise from the interaction of a known robot geometry with a known set of obstacles in the 2/3D world (workspace\textbackslash environment). 

In this paper we discuss SBMP works which attempt to replace the uniform sampling of configuration space with biased sampling, which takes advantage of some of this underlying structure. In section \ref{sec:def_guiding_space} we formalize the definition of a \emph{motion planning guiding space}, a phrase which has been used informally by many researchers in seemingly distinct contexts. Next, in section \ref{sec:prior_work}, we provide a categorization of some these past works, and indicate how they are indeed guiding spaces. Finally, in section \ref{sec:metric}, we describe an information theoretic metric which provides a more nuanced evaluation of biased sampling methods, and then demonstrate this metric on a set of algorithms in Section \ref{sec:experiments}.

We note that we are not the first to suggest alternative metrics for evaluating motion planning, for example in \cite{pearce2006metrics}, the authors describe a metric that can be used to evaluate probabilistic roadmaps (PRM), by classifying new samples based on their effect on the graph structure. Moreover prior work has used information theoretic concepts in motion planning. For example in \cite{burns2003information}, the algorithm adds configurations that maximize the expected information gain, and \cite{barraquand1995motion} attempts to compute an information preserving space based on state aggregation. In this work, our focus is on applying such ideas to guiding spaces, and as such we focus our discussion of related work in section~\ref{sec:prior_work} on algorithms that employ guiding spaces.


\section{Definition of Guiding Space}
\label{sec:def_guiding_space}

\begin{figure}
    \centering

    \begin{subfigure}[t]{0.23\textwidth}
        
        \label{fig:guiding space usage: workspace}
         \includegraphics[width=\columnwidth, page = 4]{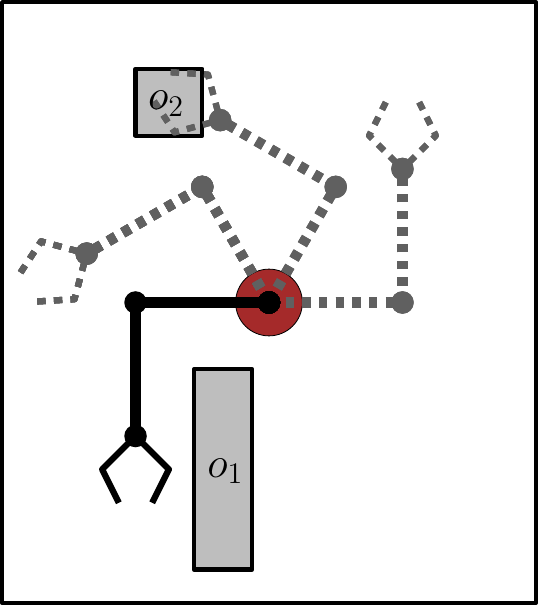}
         \subcaption{A 2D workspace with a stationary 2-joint manipulator, and 2 obstacles. Three intermediates from the path found in guiding space (b) are shown in dashed gray, where the last is the target configuration. We observe that one intermediate is invalid.}
    \end{subfigure}
    \hfill
    \begin{subfigure}[t]{0.23\textwidth}
        \label{fig:guiding space usage: c-space}
         \includegraphics[width=\columnwidth, page = 3]{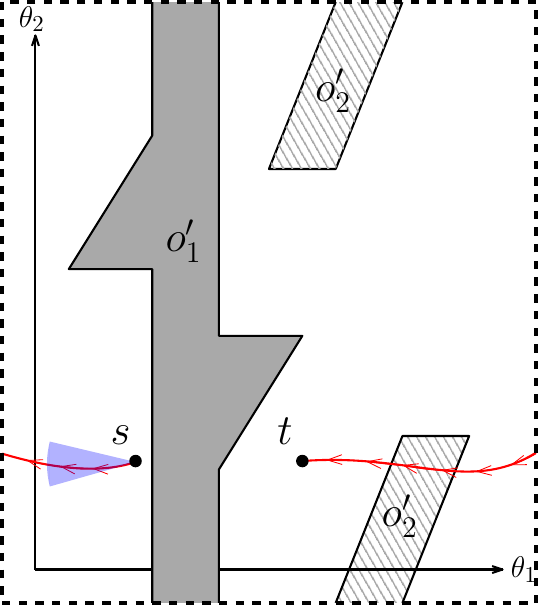}
         \subcaption{The lazy planning guiding space for the environment (a) without obstacle $o_2$. A red curve shows a path between the current and target configurations shown in (a), and a blue wedge shows the resulted sampling bias.}
    \end{subfigure}
    \caption{An illustration of guiding space usage.}
    \label{fig:guiding space illustration}
    
    \vspace{-0.6cm}
\end{figure}

\begin{definition}[Guiding Space]
Let $r$ be a robot, and $E$ be an environment. Denote the implicit configuration space given by $r$ and $E$ as $\C$. A Guiding Space is a reduction from a motion planning instance in $\C$ to a motion planning instance in another space $\GS$. It consists of a pair
\begin{equation*}
    (f: \C \longrightarrow \GS, h: \GS \times \GS \longrightarrow \Delta(\C)).
\end{equation*}
We refer to $\GS$ as the Guiding Space, $f$ as the \emph{projection} which maps instances of motion planning problems in $\C$ to $\GS$, and $h$ as the \emph{heuristic} which provides guidance (e.g., biased samples) via a distribution $\Delta(C)$ over the continuous $\C$. 
\end{definition}

This definition gives rise to a natural motion planning algorithm. Given a motion planning instance composed of start and goal configurations $(s,t)$ in $\C$, and a guiding space $(f, h)$, guided planning produces a motion planning instance $(u,v) = (f(s), f(t))$ in $\GS$, and then new samples in configuration space, $x \sim h(u,v), x \in \C$. 

Note that the algorithm can create a set $\{\GS_i\}$ of spaces, combining the guidance they provide. Alternatively it can create a sequence of such spaces, and use each guiding space $\GS_i$ to provide a single new sample in $\C$, which informs the construction of the next space $\GS_{i+1}$. The definition is general, for example we do not restrict the structure of $f$ and $h$, intentionally not limiting the set of algorithms that can be interpreted as using guiding spaces, and rather identifying the components of guidance which can be individually analyzed.

Finally we observe that the definition inherently lends itself to a recursive hierarchy, in which a guiding space $(f, h)$ can itself have a guiding space which informs how $h$ solves motion planning instances in $\GS$. Similarly, we can consider the trivial guiding spaces: $\GS = \C$ and $\GS = \{0\}$. When $\GS=\C$ we have $f(s) = s$, the identity function, and $h(s,t)$ is a constant distribution which produces an $(s,t)$ path $P_{s,t}$. This represents perfect guidance along a solution to the motion planning instance. When $\GS = \{0\}$ we have $f(x) = 0$ and $h(f(s),f(t)) = Q_\C$, where $Q_\C$ is some distribution over $\C$. This case represents sampling bias without a guiding space. In the next section we describe a set of categories of more interesting guiding spaces, and how past works fit into these categories.

\section{Analysis of previous work}
\label{sec:prior_work}
We separate guiding space methods into three main categories: robot modification, environment modification and experience based guidance. We find these subcategories natural and comprehensive, as the entire motion planning problem is encapsulated in the structure of the configuration space, which can be learned from experience, or is induced by its robot and environment constraints.




\subsection{Robot Modification}

Recalling that a configuration space is the product of a robot $r$ and environment $E$, $\C = (E, r)$, we define robot modification as producing a guiding space $\GS = (E, r')$ for some new robot $r'$. In this paradigm, the projection function $f$ describes the relationship between robot configurations, mapping configurations in the first robot's configuration space to configurations in the second. 

Given such a mapping $f$, $h$ must compute (in a possibly implicit manner) paths between configurations in $\GS$, and then use those paths to describe a sampling distribution in $\C$. Note that the translation of a path back into $\C$ is often costly and may be one-to-many (e.g., inverse-kinematics). 

There are many natural ways one might modify a robot description to obtain a closely related configuration space. One of the most natural projections is to define $\GS$ as workspace itself, ignoring the robot entirely and treating it as a fully controllable point robot in the environment \cite{holleman2000framework, DennySBA16}. Some prior works ignore kinodynamic constraints on the robot \cite{PlakuKV07}, thereby simplifying the complexity of paths in the guiding space where the robot is locally controllable. For chain-link robots, such as manipulators, a common guiding space involves focusing on a subset of the degrees of freedom \cite{lozano1987simple}. Such methods often use heuristics in deciding which subsets to consider, e.g. focus on the first joints of a manipulator. Finally, some works attempt to learn a subspace of the robot's degrees of freedom, e.g. using statistical projections such as PCA \cite{MahoneyBJ10}.

\subsection{Environment Modification}
Environment modification refers to guiding spaces which map $\C=(E, r)$ to $\GS=(E', r)$. Unlike in robot modification, the associated projection $f$ is often simpler to compute, and translation of paths in $\GS$ back to $\C$ requires relatively little work. In fact, in all of the papers cited below, $f$ is nothing but the identity function, which only changes the validity of some points in $\C$ to produce $\GS$.


Narrow passages are a known bottleneck for randomized sampling algorithms, as such, some works modify obstacles in an attempt to widen passages \cite{BayazitXA05, VonasekPK19}. Lazy planning methods \cite{BohlinK00, Hauser15} also perform a type of environment modification by ignoring constraints during initial planning, and incorporating them later as needed. When removing constraints, paths in $\GS$ are not necessarily valid in $\C$, and thus often such works iteratively or hierarchically compute a guiding space, use $h$ to produce samples near the unvalidated path, and then repeat with a new guiding space informed by the recently acquired samples. 

Unlike the above works which remove environment constraints, adding constraints to the environment is less common, but can be done to simplify geometries and therefore collision detection \cite{GhoshTMRA16}. Moreover adding constraints ensures that any path in $\GS$ is also valid in $\C$, simplifying the translation of paths found in guiding space.

\subsection{Experience Based Guidance}

Experience based guidance methods are characterized by their use of a dataset of solved motion planning problems in a diverse set of environments and/or robots. Each element of the dataset represents a guiding space, and $f$ performs a database lookup. The resulting solution must then be translated (i.e., ``fixed'') by $h$ into the current configuration space. Often such methods assume some repetitive structure, such as a known distribution of obstacles, and hope to capture this structure through appropriate feature selection to determine how $f$ queries the dataset.

The most common version of experience based guidance assumes the dataset contains a variety of environments all for the same robot. Consequently, database queries often consist of evaluating the degree to which paths in the dataset violate discovered constraints in the current motion planning instance \cite{berenson2012robot}. Some methods perform multiple queries, where the guiding space effectively becomes the composite space of multiple independent motion planning problems from the dataset. At the extreme end we have learning based methods, which train a neural network from a dataset of experience to provide guidance, effectively combining all past solutions through the weights of the neurons of a neural network \cite{bhardwaj2017learning}.

\section{Evaluation metric}
\label{sec:metric}

Often we evaluate sampling based motion planning methods using algorithm level holistic metrics such as running time or number of samples. Minimizing such metrics is a reasonable goal, but they can be sensitive to implementation details, and focusing on them can obscure which aspect of a proposed contribution is responsible for results. In the context of guided motion planning, we now present a new method to evaluate quality of the guidance.

A guiding space biases sampling, and in doing so it provides the algorithm with some information, reducing uncertainty with regards to where one should explore to connect two points in the environment. The quality of this information is exactly what we wish to capture when evaluating a guiding space. In particular, a good guiding space should contain little extra information that isn't in some desirable target distribution. 

\begin{definition}[Sampling efficiency]
Given a target sampling distribution $T$ and empirical sampling distribution $Q$, the sampling efficiency of $Q$ is defined as the information gain of using $T$ instead of $Q$, equivalently described as the Kullback-Liebler divergence between the two distributions, $$SE_T(Q) = D_{KL}(T || Q).$$
\end{definition} 



The lower this quantity, the more similar the two distributions, and therefore the more samples from $Q$ are informative about $T$. We emphasize that the choice of target distribution is application dependent. If $h$ is used to narrow down a set of possible steering options in a kinodynamic RRT, then $T$ may refer to the subset of steering options which take the agent closer to the goal. If the task is not only $(s,t)$-connectivity, but also to find an optimal path, then $T$ might refer to the set of samples on boundedly optimal paths. In the next section we describe one possible general definition of $T$ for problems in which the task is to find any path from $s$ to $t$.

\section{Experiments}
 \label{sec:experiments}

\begin{figure*}[h!]
    \begin{subfigure}[t]{0.3\textwidth}
        \includegraphics[width=\columnwidth, height=5cm]{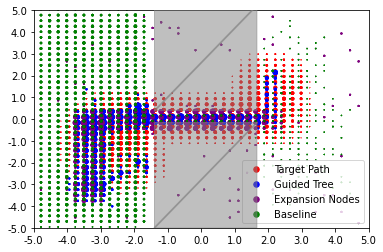}
        \caption{DR-RRT on Simple Passage. As expected, baseline RRT struggles to find the narrow passages whereas DR-RRT (with the help of workspace skeleton guidance) finds it without exploring the entire region.}
        \label{SimplePassage}
    \end{subfigure}
    \hfill
    \begin{subfigure}[t]{0.3\textwidth}
        \includegraphics[width=\columnwidth, height=5cm]{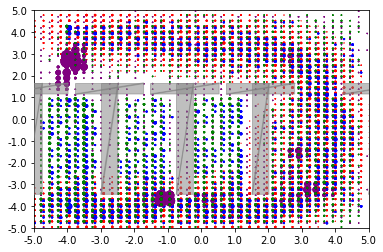}
        \caption{IRC on Trap environment. In this environment, shrinking the robot makes impassable passages seem traversable, and therefore the guidance does a poor job by biasing sampling towards those regions.}
        \label{Trap}
    \end{subfigure}
    \hfill
    \begin{subfigure}[t]{0.3\textwidth}
        \includegraphics[width=\columnwidth, height=5cm]{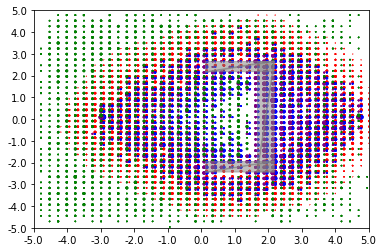}
        \caption{LazyPRM on Cup environment. Notice that in this environment there are multiple homotopy classes of solutions, and that LazyPRM, while it does oversample near the obstacles, is heavily biased along paths to the goal.}
        \label{Cup}
    \end{subfigure}
    
    \caption{We show sampling distributions for our three environments and three guiding space algorithms - (a) Simple Passage and DR-RRT, (b) Trap and IRC, (c) Cup and LazyPRM. The sampling distributions in each figure correspond to an average over 15 random seeds, and are displayed as a discretized grid for visual clarity, where the size of each circle corresponds to the number of samples in that grid cell. In red we show the approximate distribution of the target path $T_P^\delta$. In blue are the set of samples on the tree resulting from guided samples. In purple are the set of expansion samples, i.e., the samples in the environment towards which the tree is steered. Finally, in green are the samples of a baseline RRT.}
\end{figure*}

Given a path $P$ we define a target distribution $T_P^\delta$ as a uniform distribution over points at most a distance $\delta$ from the path, $T_P^\delta = U(\{x \mid d(x, P) < \delta\})$. This represents the optimal sampling distribution for discovering the path $P$. Let $Q$ be the sampling distribution induced by a guiding space, then the guiding space sampling efficiency is $SE_{T_P^\delta}(Q) = D_{KL}(T_P^\delta || Q) = \mathbb{E}_{T_P^\delta}(\frac{\log T_P^\delta}{\log Q})$.

We evaluate the sampling efficiency of four methods on three environments, designed to highlight the strengths and weaknesses of the guidance provided by each.
Dynamic Region-biased RRT (DR-RRT)~\cite{DennySBA16} grows an RRT while constraining sampling around a topological skeleton of the workspace environment.
LazyPRM~\cite{BohlinK00} does not perform collision detection while extending the edges of the tree, and rather discards invalid edges when later evaluating paths. 
Iterative Relaxation of Constraints (IRC)~\cite{BayazitXA05} finds a path for a shrunken version of the robot, effectively widening narrow passages, then uses this path as guidance for the full-size robot.
RRT~\cite{lavalle1998rapidly} is used as a baseline which uses no guidance.
All methods were implemented in C++ using the Parasol Planning Library. 

All scenarios consider a 3 degree of freedom (2 translational, one rotational) robot with dimensions 0.2x0.6 in a 10x10 2D environment.
\textit{Simple Passage} (Fig.~\ref{SimplePassage}) is composed of a narrow passage (width 0.5) that the robot must pass through in a lengthwise orientation to reach the goal.
\textit{Trap} (Fig.~\ref{Trap}) includes several false passages that the robot may attempt to pass through before finding the feasible passage to the goal.
\textit{Cup} (Fig.~\ref{Cup}) features a large obstacle obstructing a direct path to the goal.
In all cases we set $\delta=0.5$, corresponding to the RRT maximum steering distance, meaning the maximum distance a new sample is extended from the tree. We make this choice of $\delta$ because it relates to the natural density of samples in the environment, though we note there are other reasonable choices, such as the maximum clearance of $P$.

We report results in Table~\ref{table}, where for each combination of algorithm and environment we report the KL-divergence derived sampling efficiency of the algorithm relative to the target distribution $T_P^\delta$ for the resulting $(s,t)$-path $P$. Notice that guidance is not always beneficial: the environment Trap was designed as an adversarial example for methods like DR-RRT and IRC, and indeed they perform similarly or worse than baseline RRT in this environment. While Cup was designed to hinder LazyPRM, it seems that lazy planning does quite well under our metric. 

\begin{table}[h!]
    \centering
    \begin{tabular}{|c|c|c|c|}
        \hline
        & Simple Passage & Trap & Cup  \\
        \hline
        \rowcolor[HTML]{EFEFEF}
        RRT &
        \begin{tabular}{r} 32.147 \\\hline 18.671  \end{tabular} &
        \begin{tabular}{r} 9.589 \\\hline 7.547  \end{tabular} & 
        \begin{tabular}{r} 8.539 \\\hline 5.246  \end{tabular} \\
        \hline
        LazyPRM &
        \begin{tabular}{c} 11.395 \\\hline NA  \end{tabular} &
        \begin{tabular}{c} 14.869 \\\hline NA  \end{tabular} & 
        \begin{tabular}{c} 6.992 \\\hline NA  \end{tabular} \\
        \hline
        \rowcolor[HTML]{EFEFEF}
        DR-RRT &
        \begin{tabular}{r} 6.115 \\\hline 19.845  \end{tabular} &
        \begin{tabular}{r} 38.759 \\\hline 57.244  \end{tabular} & 
        \begin{tabular}{r} 5.834 \\\hline 18.176  \end{tabular} \\
        \hline
        IRC &
        \begin{tabular}{r} 13.797 \\\hline 19.447  \end{tabular} &
        \begin{tabular}{r} 4.845 \\\hline 21.612  \end{tabular} & 
        \begin{tabular}{r} 3.829 \\\hline 9.443  \end{tabular} \\
        \hline
    \end{tabular}
    \caption{We report the KL-divergence values $SE_T(Q)$, where $T = T_P^\delta$, for final path $P$ found by the algorithm, and $Q$ is the distribution produced by the algorithm in the same row. In each cell the upper value represents the samples on the final RRT tree, and the lower value represents the samples used to guide the expansion of nodes in the tree.}
    \label{table}
\end{table}

\section{Discussion}
In this work we introduced the concept of a guiding space for motion planning. This definition captures the common paradigm in which one solves ``simpler'' versions of motion planning instances in order to get guidance in the form of biased planning for an original motion planning problem.

Our characterization is useful because it immediately follows that one can evaluate quality of guidance separately from implementation dependent metrics such as overall algorithm running time. Similarly, the metric we propose for evaluating a guiding space isolates the \emph{information} contained within it regarding the underlying exploration task, rather than how that information is used. A better understanding of what information is captured by a guiding space can consequently inform how and when we use it. 

As the complexity of motion planning algorithms increases, and as black-box experience based methods become ubiquitous, it is especially important to have evaluation metrics that are informative regardless of lower level details. Simultaneously, it is crucial to recognize which methods are indeed comparable. The framework of guiding spaces demonstrates that a large class of motion planning algorithms are more similar than they first appear.

\bibliography{refs}
\end{document}